\documentclass[10pt, twocolumn]{article}

\usepackage[utf8]{inputenc}
\usepackage{multirow}
\usepackage{amsmath}
\usepackage{graphicx}
\usepackage{subfig} 
\usepackage{caption}

\usepackage{soul}
\usepackage{xcolor}

\usepackage{geometry}
\begin{document}
\title{Deepzzle: Solving Visual Jigsaw Puzzles with Deep Learning and Shortest Path Optimization}
\author{Marie-Morgane~Paumard,
        David~Picard
        and~Hedi~Tabia}
\date{}

\maketitle

\begin{abstract}
We tackle the image reassembly problem with wide space between the fragments, in such a way that the patterns and colors continuity is mostly unusable. The spacing emulates the erosion of which the archaeological fragments suffer. We crop-square the fragments borders to compel our algorithm to learn from the content of the fragments. We also complicate the image reassembly by removing fragments and adding pieces from other sources. We use a two-step method to obtain the reassemblies: 1) a neural network predicts the positions of the fragments despite the gaps between them; 2) a graph that leads to the best reassemblies is made from these predictions. In this paper, we notably investigate the effect of branch-cut in the graph of reassemblies. We also provide a comparison with the literature, solve complex images reassemblies, explore at length the dataset, and propose a new metric that suits its specificities.

\end{abstract}

\section{Introduction}
\label{sec:introduction}

From cultural heritage to genome biology \cite{hashem}, numerous problems revolve around our ability to perform automatic reassemblies. In the case of archaeology, museum collections regroup a large amount of mixed 2D or 3D fragments of art masterpieces. Finding the correct reassemblies is a crucial step to understand our past better. Usually, this task relies on computer vision algorithms, such as contours or features detection \cite{vellaichamy}. The recent upsurge of deep-learning opens bright perspectives for finding better reassemblies more efficiently.

\begin{figure}[htb]
    \subfloat{\includegraphics[width=0.45\linewidth]{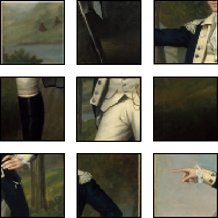}}\hfill
    \subfloat{\includegraphics[width=0.45\linewidth]{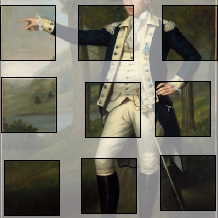}}
    \caption{A $3\times 3$ jigsaw puzzle-solving.}
    \label{fig:task}
\end{figure}

In \cite{paumard1}, we proposed a preliminary method to tackle the puzzle-solving task with deep neural networks and graphs. We focused on solving $3\times 3$ jigsaw puzzles made of same-sized squared 2D fragments (Figure \ref{fig:task}), using a 2-step method. First, given a central fragment, we used a neural network to predict the relative position of each remaining fragment. Then, the best solution is obtained using a graph of the possible reassembly.

In this paper, we propose an extension of \cite{paumard1}, which is called Deepzzle. Our contributions are the following. We reframe the puzzle-solving task with a probabilistic model. We improve the graphs used for the reassembly step, in such a way we can select the fragments from a bigger ensemble. We propose a robust metric to evaluate the visual quality of the reassembly better. We adapt our method to compare it with the literature. We solve tougher puzzles made from the combination of the main variations explored in \cite{paumard1}, such as missing pieces, unknown central fragment, additional unrelated fragments or fragments from different photographs of the same object. We also propose a comprehensive analysis of the MET dataset.

This paper is organized as follow: in Section \ref{sec:soa}, we review the literature on archaeological puzzles, and we complete this introduction by giving an overview of puzzle-solving task based on deep learning. Then, we detail our method in Section \ref{sec:mth}. In Section \ref{sec:exp}, we present the dataset and introduce a new metric. Then, we explore the effects of the branch-cuts in the graph, and we delve into the reassemblies with missing and outsiders fragments.

\section{State of the Art}
\label{sec:soa}
\subsection{Archaeological puzzles}
Aside from the jigsaw puzzle solving itself, which is an NP-complete problem, the solving of archaeological puzzles requires to address a variety of tricky issues. Examples are the non-square shape of the fragments, the very different size of the fragments, the erosion of the fragments contours and colors, the missing fragments, the mixture of fragments from different objects and the continuity of the space of the relative transformations between a couple of fragment.

The case of automatic reassembly is extensively studied for the restoration of cultural sites and objects, as Rasheed and Nordin highlight in their surveys \cite{rasheed1, rasheed2}. Each work they cited focus on one, two or sometimes even several issues related to archaeological puzzles (such as a large number of pieces, missing fragments, unknown orientation, erosion, fading colors). We can divide most of the methods into two categories depending on whether the aim is a coarse positioning of the fragments and a precise reassembly.

Most of the state-of-the-art methods exploit either the fragments global shape or their contours \cite{son2, huang, papaioannou, mcbride, liu, zhu, zhang2}, while some focus on the content (such as colors or patterns) of each fragment \cite{gur, paikin, sholomon, gallagher, son1}. Usually, the approach by content uses square fragments and focus on the puzzle-solving task itself without any archaeological ambition. For example, Son et al. \cite{son1} present an algorithm to solve square jigsaw puzzles based on a pairwise matching of the colors and patterns of the fragments. Their method is efficient on very large puzzles (several thousand of pieces). It can reassemble effectively two mixed puzzles. Similarly, Paikin and Tal \cite{paikin} solve large jigsaw puzzles. They focus on the missing fragments and obtain excellent reassembly in that case.

Conversely, solving by fragments shape prioritizes the variety of shapes and the 3D puzzles over the large puzzles with missing fragments cases. The methods usually start from a study of fractured surfaces, as they are the one to be reassembled. For example, Zhang et al. \cite{zhang2} address the issue of reassembling 3D broken artifacts by discovering the fractured regions and trying to match them. When available, they use a general template of the artifact to place the fragments approximatively. Their algorithm performs well on small pieces but does not take into account the erosion. Just as in \cite{paumard1}, they use a graph model to find the best reassemblies.

Papaioannou et al. \cite{papaioannou} also propose a set of tools for semi-automated reassembly of 3D archaeological objects based on the fragment content. They present a pipeline for both the puzzle-solving and the completion tasks. First, they separate fractured surfaces from the potentially fractured and the entire surfaces. To counter the effect of erosion or material damages, the user may append information to the unusable surfaces. The next step is the computation of the pairwise scores of fractured regions. Then, the authors use a combinatorial solver to select the best matches and build the reassembly.

Finally, some techniques exploit both the fragments shapes and content \cite{zhang1, derech}. Zhang and Li \cite{zhang1} introduce a method based on both fragment shapes and patterns. They use shapes to propose matching between fragments and evaluate the matching with the borders and the colors of the fragments. 

In \cite{derech}, the authors propose to examine the overlapping of extrapolated fragments rather than searching for valid continuations. They solve the puzzle one piece after another: they use the current reassembly to place the next fragment. Their algorithm is efficient at tackling almost all of the diverse concerns of the archaeological puzzles with the  of numerous missing fragments. They also consider a slight erosion of the fragments borders and tackle it by using inpainting techniques.

As deep learning brought efficient solutions in various computer vision tasks, we expect that the archaeological puzzles tasks benefit from deep learning.

\subsection{Puzzle solving with deep learning}

Independently of the reassembly as a goal, the jigsaw puzzle-solving task is commonly used to discover visual features in an unsupervised learning setup. Here, it is no longer a matter of finding a precise reassembly: the goal is to propose a coarse positioning of the fragments.

Doersch et al. \cite{doersch} pioneered this topic by proposing an architecture to solve $3 \times 3$ square-puzzles. Given a central fragment, they predict the relative position of any adjacent fragment. Afterward, they use the newly learned features on a wide variety of vision tasks with success. Paumard et al. \cite{paumard1, paumard2} improve their method and propose a puzzle-solving based on the relative position prediction. They propose a few variations on the $3 \times 3$ problem, such as the case of the missing fragments. In this paper, we propose an extension of this work.

Other work \cite{noroozi1, noroozi2, wei, santacruz, kim} also study the jigsaw puzzle as a pretext task. Noroozi and Favaro \cite{noroozi1} solve $3 \times 3$ puzzles with only a neural network: it receives all the 9 pieces as an input and predicts the correct fragments permutation. As their network requires 9 fragments, they cannot solve puzzles with missing or outsider fragments. Moreover, due to the high number of permutations  ($\geq10^5$), that are the classes of the network, they face tremendous computation time. They avoid this issue by restricting the number of possible reassemblies, which causes most of them to be unattainable.
While preserving their architecture, they complicate the resolution task in \cite{noroozi2} by replacing 1 or 2 fragments of the puzzle by fragments extracted from a random image. This setup is not equivalent to solving puzzles with 2 missing and 2 outsider fragment, as the two outsider fragments cannot be labeled as outsiders.

Wei et al. \cite{wei} propose an iterative method to solve bigger and 3D-puzzles. They combine two predictions: one from the pairwise relative position as \cite{doersch} and one from the absolute position in the puzzle. This last position is predicted based on all the other fragments, as in \cite{noroozi1}. Their architecture may accommodate with missing and outsider fragments, which make their method close to ours.

In \cite{santacruz}, the authors propose an architecture that can reorder images sequences. Given a set of faces, they can, for instance, order them by the expressions. They claim their neural network enable to solve $3\times 3$ jigsaw puzzles. Using the reordering as a pretext task, their architecture achieves better results than \cite{doersch, noroozi1} on the usual classification tasks. However, they do not evaluate their architecture on the puzzle-solving task.

Kim et al. \cite{kim} tackle the case of decoloration with one missing fragment. Based on inpainting and colorization techniques, they fully restore the images. To do the reassembly, they use a network similar to Noroozi et al. \cite{noroozi1} in which they input a white fragment, representing the missing tile. The authors do not provide insight into the effectiveness of their architecture for the reassembly.

In this paper, we are interested in solving the jigsaw puzzle per se and not in learning generic visual features. Thus, our method searches for the most probable reassembly among all possible reassemblies.

\section{Methods}
\label{sec:mth}
The archaeological puzzles show some properties that make them especially worthwhile to address. They are rarely complete, and the fragments come in various sizes and shape. They also suffer from erosion and color fading. In this paper, we suppose that the erosion significantly damaged the pieces, making the fragments borders unusable. As we do not take the borders into account, we can only predict a coarse position. The precise puzzle solving is then to be done from the results of our coarse reassembly. Nevertheless, an algorithm based mainly on the content of the fragments is more resistant to missing pieces, as it can handle large gaps in the reconstructions.

In this paper, the space between two fragments is about the same as the half-side of a fragment. We choose such a tremendous value for mimicking the erosion because it is equivalent to crop-square any piece that is almost square. In sum, we solve jigsaw puzzles characterized by a high rate of missing fragments, a single size of fragments and a considerable erosion. To a certain extent, the erosion can conceal a various set of fragment shape. We also cover the case of a puzzle mixed with fragments from other puzzles and puzzles where the central fragment is not known.

After providing an overview of the Deepzzle method, we formulate the problem with a probabilistic model. Then, we use this framing to predict the positions of the lateral fragments. Finally, we detail the reassembly based on these predictions.

\subsection{Method overview}
\begin{figure*}
    \centering
    \includegraphics[width=\linewidth]{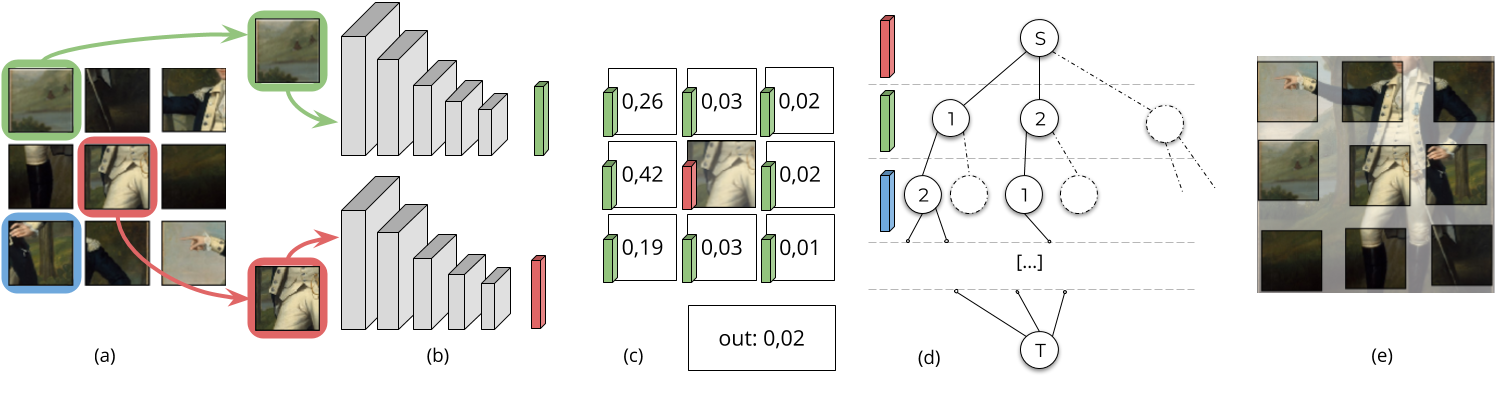}
    \caption{Outline of the Deepzzle method. From a set of pieces (a) made of a central fragment (in red) and lateral fragments, we pick a lateral fragment (in green). We extract its features (b) and predict its place among the eight lateral positions and the outsider class (c). Then, we build the graph of the prediction (d) in which each line matches with a fragment. The reassembly (e) is computed from the shortest path in the graph.}
    \label{fig:overview}
\end{figure*}

Figure \ref{fig:overview} illustrates our puzzle-solving process. For each image in our dataset, we extract a square that we cut into 9 pieces. To mimic the erosion, we then randomly crop a fragment inside each piece, making sure there is a wide gap between the fragments. Then, we pair the central fragment with each lateral fragment. Each couple is processed by a neural network that predicts their relative position among the 8 alternatives. These probabilities are used to build a graph, in which we compute the shortest path to reassemble the puzzle.

We propose two extensions of this problem. First, we consider the case where the central fragment is unknown. In this case, we compute the relative positions supposing that each fragment is the central one. Then, we apply the shortest path algorithm in each of these graphs, and we select the most probable solution. Second, we deal with missing fragments and outsider fragments, which are frequent in archaeology. In this case, we allow fragments to be unused and positions to be unfilled.

\subsection{Problem formulation}
We are looking for the most probable reassembly: the reassembly that satisfies as many relative position predictions as possible.

We introduce $P_r$ a probability and $x_{i,j}$ the affectation of the fragment $i \in [0\mathrel{{.}\,{.}}\nobreak f]$ at the position $j \in [0\mathrel{{.}\,{.}}\nobreak 9]$, where $f+1$ is the number of fragments.  We use position 9 to label the outsider fragments. We define position 0 as the central position and fragment 0 as the central fragment. We then introduce $x_c = x_{0,0}$, the placement of the central fragment at the central position.

We want to find the maximum joint probability of placing all fragments:
\begin{equation*}
    \label{eq00}
    \max P_r(x_c, x_{1,1}, x_{1,2}, \ldots, x_{2,j_1}, \ldots, x_{f,9}).
\end{equation*}
Because each fragment can occupy only one position,
we simplify the latter equations and introduce $x_i$ the chosen affectation of the fragment $i$:
\begin{equation}
    \label{eq01}
    \max P_r(x_c, x_1, x_2, \ldots, x_f).
\end{equation}
 As the predictions of the positions of the lateral fragments depend on the central fragment, we want to extract the central fragment $x_c$ from $P_r$. We use Bayes rule:

\begin{equation*}
    P_r(x_c, x_1, \ldots, x_f) = P_r(x_1 \ldots x_f | x_c) \times P_r(x_c).
\end{equation*}

We assume $P_r(x_c) = 1$. To ease the notation, we drop the term $|x_c$ in the further equations while keeping in mind that $x_c$ conditions all probabilities .

We now restate the previous equation with Bayes rule, to expose that assembling the puzzle is an iterative process where fragments are selected and placed sequentially. As such, the probability of a reassembly depends on the probabilities of placing the last fragment, knowing that all previous fragments are placed:

\begin{multline}
\label{eq02}
P_r(x_f\ldots x_1) = P_r(x_f | x_{f-1}\ldots x_1) \times P_r(x_{f-1}\ldots x_1).
\end{multline}

To obtain a tractable approximation, we suppose that $x_i$ follows the Markov Chain:

\begin{equation}
\label{eq03}
P_r(x_f|x_{f-1}\ldots x_1) = P_r(x_f | x_{f-1}).
\end{equation}

Unrolling the recursion of Equation \ref{eq02} leads to:

\begin{equation*}
P_r(x_1\ldots x_f) = \prod_{i \in [2 \mathrel{{.}\,{.}}\nobreak f]} P_r(x_i|x_{i-1})\times P_r(x_1).
\end{equation*}

To further simplify the problem, we make the approximation that $x_i$ and $x_{i-1}$ are independent:

\begin{equation}
    \label{eq04}
    P_r(x_i|x_{i-1}) = P_r(x_{i}),
\end{equation}

which leads to:

\begin{equation*}
P_r(x_1 \ldots x_f) = \prod_{i \in [1 \mathrel{{.}\,{.}}\nobreak f]} (P_r(x_i)).
\end{equation*}

This approximation allows using the pairwise relationships to solve a puzzle. Without this approximation, the neural network architecture
would be significantly more complex as it would require to compare all the fragments. Such architecture would be less adaptable to missing and outsider fragments.

In turns, it means we want to solve the following optimization problem:

\begin{equation}
\label{eq05b}
\max P_r(x_1, \ldots x_f) = \max \prod_i P_r (x_i),
\end{equation}
which is equivalent to:

\begin{equation}
\label{eq05}
\max \log P_r(x_1, \ldots x_f) = \max \sum_i \log P_r(x_{i}).
\end{equation}

\subsection{Prediction of the relative position}

In order to solve the optimization problem of Equation \ref{eq05}, we need an estimator of $P_r(x_{i}|x_c)$. We propose to cast the problem of estimating $P_r(x_{i}|x_c)$ as a classification problem that can easily be solved by a deep convolutional neural network. The neural network has two inputs, corresponding to the central fragment and the lateral fragment, and 9 outputs corresponding to the possible positions of the fragment $i$. To optimize this network, we use a categorical cross-entropy. Remark also that the architecture we propose is directly derived from the independence approximation made in Equation \ref{eq04}.

More specifically, each fragment goes through a Siamese network (Figure \ref{fig:overview}b) that performs the same features extraction, thanks to shared weights. These Feature Extraction Networks (FEN) are described and justified in Table 1 of Paumard et al. \cite{paumard1}. 
Briefly, the two of them are fed with a fragment of size $96\times 96 \times 3$. They are made of five convolution layers followed by a fully connected layer of size 512. Then, a Kronecker product merges the features of the fragments in the Combination Layer (CL). Finally, three fully-connected (FC) layers followed by a batch-normalization and an activation (ReLU for the first two, and softmax to ensure probabilities for the last layer) predicts the relative position. We set the output size to 9, the number of free positions plus the outsider class.

\subsection{Graph-based reassembly}
\subsubsection{Building the graph}
To solve Equation \ref{eq05} with the probabilities predicted by the deep neural network, we select an arbitrary order in which we process the fragments, because the chosen order has no impact on the solution we obtain. For example, it is equivalent to place the fragment $1$ in position $1$ then the fragment $2$ in position $2$, and to place first the fragment $2$ in position $2$ then the fragment $1$ in position $1$.


We then build the graph of all the possible reassemblies with a recursive algorithm. Starting from an empty puzzle $S$, we decide where to place the first fragment $i$. We model this decision by $9$ nodes connected to $S$. The negative logarithm of the classification scores weight the edges. Then, each node is connected to the remaining positions that can be attributed to the second fragment, and so on. The last fragment is placed at the last remaining position, and it is connected to the end of the graph $T$. These last edges are given a null weight. In other words, the depth of the graph corresponds to fragments, and the width is the available positions.

\begin{figure*}[htb]
    \centering
    \subfloat[Graph with potential outsiders and 6+ missing fragments.]{\label{fig:graph-out}{\includegraphics[width=0.3\textwidth]{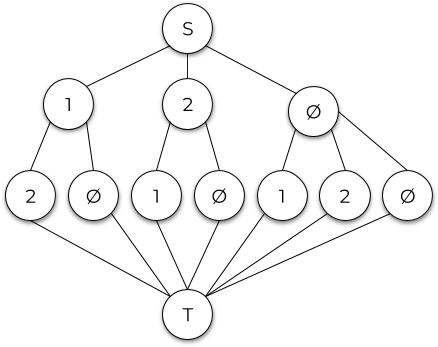}}}\hfill
    \subfloat[Graph with a cut of the fragment C for positions 1 and 2, without reordering.]{\label{fig:graph-cut-without}{\includegraphics[width=0.3\textwidth]{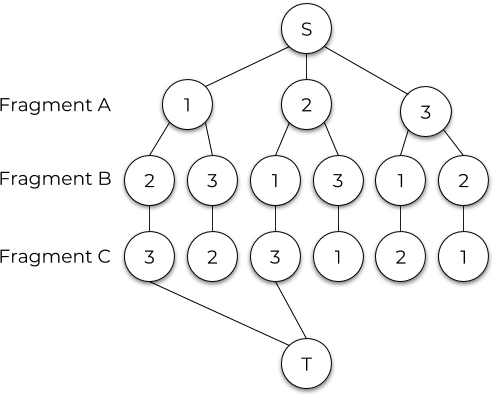}}}\hfill
    \subfloat[Graph with a cut of the fragment C for positions 1 and 2, with reordering.]{\label{fig:graph-cut-with}{\includegraphics[width=0.3\textwidth]{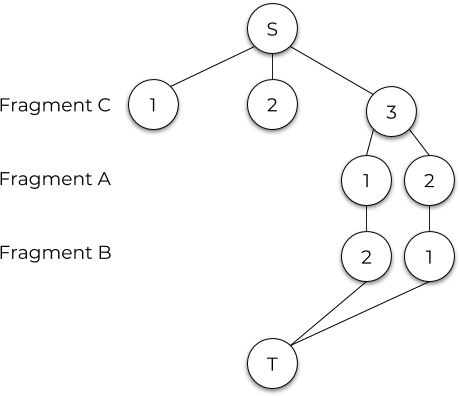}}}
    \caption{Some of the graphs we use for the reassembly.}
    \label{fig:graphs}
\end{figure*}

In the case where the central fragment is unknown, the first decision from $S$ is to select the central fragment. In the case where there are outsider fragments, an empty position node is added to each placement decision (Figure \ref{fig:graph-out}).

In order to find the most likely reassembly, we compute the shortest path from $S$ to $T$ to minimize the sum of the weights between visited nodes, which corresponds to the solution of Equation \ref{eq05}.

\subsubsection{Cuts in the graph}

The number of nodes $N$ in the graph when outsiders are allowed is obtained recursively in the following equation. We have: $N = n(f,p)+1$, where $f$ is the number of lateral fragments and $p$ is the number of available positions (maximum 8).
\begin{equation}
    \left\{
        \begin{aligned}
            n(f,p) &= p \cdot n(f-1, p-1)+n(f-1,p)+1, \\
            n(1,p) &= p + 2, \\
            n(f,0) &= f+1.
        \end{aligned}
    \right.
\end{equation}

Similarly, we get the number of edges $E = e(f,p)-1$ from the following equation:
\begin{equation}
  \left\{
      \begin{aligned}
            e(f,p) &= p \cdot e(f-i,p-1) + e(f-1,p) +1, \\
            e(1,p) &= 2 \cdot p + 3, \\
            e(f,0) &= f + 2.
      \end{aligned}
    \right.
\end{equation}

We note that $N$ is bounded from below by the number of nodes in the penultimate row. This row length corresponds to the possible reassemblies and is bounded from below by the reassemblies without any outsiders.

If $f \leq p$, then the number of reassemblies without enabling the outsiders is bounded from below by:
\begin{equation}
\label{eq_cut0}
\frac{p!}{(p-f)!}.
\end{equation}
If $f > p$, we can place $f+1$ fragments ($f$ fragments plus one empty fragment) in the first position, $f$ fragments in the second position, and we continue until the last position is filled. Thus, the number of reassemblies considering the outsiders is bounded from below by: \begin{equation}
\label{eq_cut1}
\frac{(f+1)!}{(f+1-p)!}.
\end{equation}


To tackle this complexity, we cut the branches that display a weight lower than a threshold $\theta$. Such branches correspond to a low placement probability, which in turns produces a low reassembly probability due to the multiplicative property of Equation \ref{eq05b}. Cutting enables us to improve our computation time significantly, and thus the number of outsider fragments we can take into account. If the value of a relative position prediction comes under a specific threshold, the branch is not connected to the trunk T (see Figure \ref{fig:graph-cut-without}).

As the shortest path starts from the trunk $T$ and not from $S$, the graphs on Figures \ref{fig:graph-cut-without} and \ref{fig:graph-cut-with} are equivalent. However, the latest is quicker to build, as it is smaller than the others. Thus, the sooner the cuts occur, the better it is. This observation leads to a reordering of the graph rows: the first fragments we place are these that allow the most of cuts. Remark that, although this reordering affects the size of the graph, it does not affect the number of explored reassemblies.

\section{Experiments}
\label{sec:exp}

We begin this section by presenting the dataset and the evaluation metrics. We follow by selecting the branch-cut threshold $\theta$. We provide a few baselines by comparing our method to the literature. Then, we present the reassembly case with missing and additional fragments, starting with a quantitative analysis followed by a qualitative study. Following that, we discuss the case with an unknown central fragment. We discuss the results with regards to the different classes of the dataset. Finally, we test the robustness of our algorithm with puzzles made of different photographs of the fragments.

\subsection{Dataset and metrics}
\subsubsection{Dataset}
We use the MET dataset introduced in \cite{paumard2}. This dataset provides images that have been taken with ultra-high-resolution cameras \cite{met-interview} and that avoid the lens bias that comes with the popular dataset \cite{doersch}. We use a model trained on 10000 images to predict the resolution of 2000 images. We prepare the fragments following the procedure exposed in \cite{doersch}. From a square image randomly cropped from a picture of art pieces, we extract 9 fragments of $96 \times 96$ pixels. We set the margin between the fragments to 48 pixels to simulate the erosion.

The pictures of the dataset fall into three categories, similar in size: artifacts, engravings and texts, and paintings. An artifact may be a piece of clothing, a piece of tableware, a pottery plate, a carved flint or a sculpture. As the artifact pictures display a uniform background, the background fragments are expected to be misplaced. The paintings are mostly portraits and landscapes. The engravings include of geometric engravings (around 20\% of the dataset), illustrated engravings (13\% of the dataset) and printed texts (less than 1\% of the dataset). A fifth of the dataset is composed of black and white images.

\subsubsection{Metrics}
\begin{figure}
    \centering
    \includegraphics[width=\linewidth]{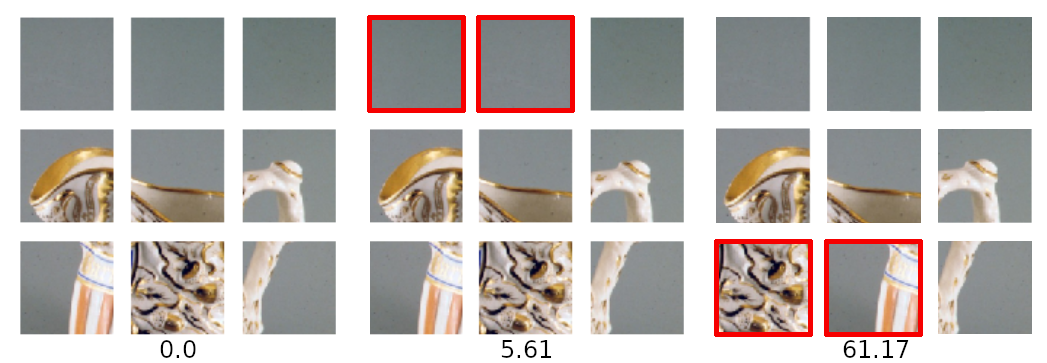}
    \caption{Selection of the best threshold for the third metric. The red outline shows the fragments that are misplaced. The case described by the third image is typical: the upper fragments are so similar that they are swapped. The values below the reassemblies are the difference between the prediction and the solution.}
    \label{fig:metric3}
\end{figure}

To assess the quality of the reassembly, we use three metrics. The first one computes the number of correct reassemblies. The second one calculates the number of well-placed fragments. However, in numerous images of the dataset, we have few indistinguishable background fragments (see Figure \ref{fig:metric3}) which lead to a random prediction that scores poorly with the previous metrics. We look for a visually plausible solution rather than the exact one. For example, some archaeological puzzles contain similar fragments that can often be swapped, e.g., the limestone blocks of a Roman temple. We consider as successful any reassembly where similar fragments are swapped. Then, we introduce a third metric that reflects this objective of visually acceptable reassembly.
It evaluates the number of almost correct reassemblies by measuring the similarity between fragments. When two similar fragments are swapped, the puzzle is still considered correctly reassembled if the norm of the difference between the fragment of the solution and the fragment of the predicted reassembly is below a threshold.

In Figure \ref{fig:metric3}, we show an example of the values of the threshold based on the fragments that are misplaced. We performed statistical analysis and set the threshold to 20, as this value confuses most of the similar fragments without allowing wrong switches.

\subsection{Branch-cut evaluation}
\begin{figure}[!t]
    \centering
    \includegraphics[width=\linewidth]{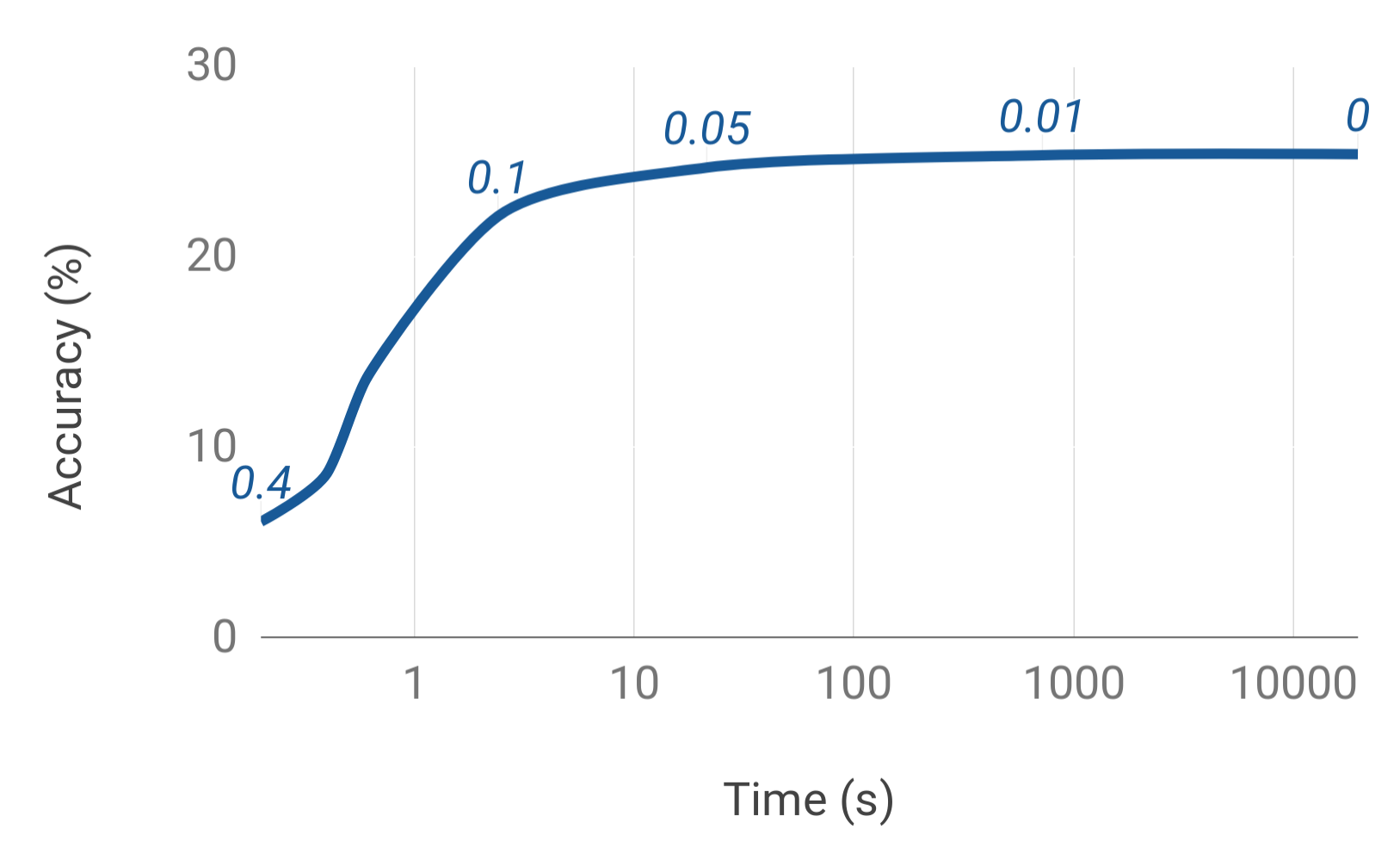}
    \caption{Comparison of the reassembly time for various cut values.}
    \label{fig:graphcut}
\end{figure}

We evaluate the trade-off between  accuracy and computational time for different values of the threshold in our branch cut strategy in Figure \ref{fig:graphcut}. As a baseline, solving a full $3 \times 3$ puzzle takes about 20,000 s. Setting the threshold $\theta$ to 0.01 allows us to gain an order of magnitude without any loss of accuracy. Setting $\theta$ to 0.05 leads to a gain of 3 orders of magnitude, or about 20s per reassembly, with a marginal loss of accuracy. We consequently use a threshold of 0.05 in the remaining experiments.

\subsection{Reassembly baselines}

We provide three different ways to assess the effectiveness of our method, in comparison with the literature. We analyze the ability of our neural network to predict the relative position, the final reassembly correctness, and the results on a larger and more popular dataset.

Doersch et al. \cite{doersch} proposed a method that allows predicting the relative position of lateral fragments, among 8 classes. We replicate their architecture and obtain 57\% of accuracy. Our 8-classes architecture is inspired by their work and had been modified to reach better accuracy. We achieve 65\% of accuracy on the same problem. 

Noroozi and Favaro \cite{noroozi1} proposed an end-to-end method: they take 9 fragments as input and learn the correct permutation among a limited number of arbitrarily chosen permutations. We reproduce their setup and apply it to the MET dataset for 10, 100, and 1000 permutations. We use our architecture to extract the features of each fragment, i.e., before the feature merging. To compare with our method, we cut the tree in such a fashion that the possible paths correspond to the allowed permutations. We use our 8-classes network, and we use graph solving for unknown central fragment. The results are exposed in Table \ref{tabl:noroozi}.

\begin{table}[h]
    \renewcommand{\arraystretch}{1.3}
    \centering
    \begin{tabular}{|l|c|c|c|c|}
        \hline
        & \multicolumn{4}{c|}{Number of permutations} \\
        & 10 & 100 & 1000 & 9! \\
        \hline
        Favaro and Noroozi [22] & 86.6  & 69.3 & 51.6 & - \\
        Ours with unknown center & 91.5 & 81.7 & 64.8 & 39.2 \\
        \hline
    \end{tabular}
    \caption{Percentage of correct image reassembly.}
    \label{tabl:noroozi}
\end{table}

We observe that our process greatly surpasses Favaro and Noroozi’s in reassembly scores. As we apply pairwise comparison on the input fragments, we can see it as a subtask of the permutation classification. In one respect, we better guide the learning process. Moreover, we recall that our method offers two other benefits over Favaro and Noroozi’s: it covers all the possible permutations, and it handles outsider fragments.

Last, we solve puzzles from the ImageNet validation dataset. We use our 9-classes network trained on the MET dataset only. We choose not to retrain our deep neural network, to avoid the lens bias that comes with the popular datasets \cite{doersch}. We obtain 78.6\% of well-placed fragments and 48.5\% of perfect reassemblies.

\subsection{Reassembly with missing and additional fragments}

\subsubsection{Quantitative reassemblies}
\begin{table}[!t]
\renewcommand{\arraystretch}{1.3}
\centering
\begin{tabular}{|p{15mm} c|c|c|c|c|}
\hline 
\multicolumn{2}{|c}{\multirow{2}{*}{}} & \multicolumn{4}{|c|}{Number of outsiders} \\
&& 0 & 1 & 2 & 3\\
\hline
\multicolumn{2}{|c|}{\shortstack{Benchmark: \\ perfect reassemblies}} & 22.1 & 18.4 & 16.8 & 15.4 \\
\hline
\multirow{8}{*}{\shortstack{Number of \\ missing \\ fragments}} & 0 & 24.7 & 19.9 & 18.3 & 16.9\\
& 1 & 20.8 & 12.9 & 11.3 & 11.0 \\
& 2 & 21.1 & 10.6 & 8.8 & 8.3 \\
& 3 & 22.6 & 12.0 & 9.8 & 6.5 \\
& 4 & 24.9 & 12.2 & 8.4 & 6.8 \\
& 5 & 31.1 & 16.6 & 10.9 & 8.3 \\
& 6 & 43.4 & 22.7 & 13.9 & 10.6 \\
& 7 & 64.0 & 33.7 & 21.0 & 13.0 \\
\hline
\end{tabular}
\caption{Percentage of almost-perfect reassemblies of the images, with a cut of 0.05.}
\label{tabl:reassembly_images}
\end{table}

\begin{table}[!t]
\renewcommand{\arraystretch}{1.3}
\centering
\begin{tabular}{|p{15mm} c|c|c|c|c|}
\hline 
\multicolumn{2}{|c}{\multirow{2}{*}{}} & \multicolumn{4}{|c|}{Number of outsiders} \\
&& 0 & 1 & 2 & 3\\
\hline
\multirow{8}{*}{\shortstack{Number of \\ missing \\ fragments}} & 0 & 64.6 & 62.8 & 60.9 & 60.3 \\
& 1 & 61.6 & 59.4 & 57.9 & 57.0 \\
& 2 & 61.1 & 57.8 & 55.7 & 55.4 \\
& 3 & 63.0 & 59.6 & 57.3 & 54.6 \\
& 4 & 66.9 & 62.0 & 58.3 & 56.0 \\
& 5 & 72.4 & 66.9 & 62.2 & 59.3 \\
& 6 & 80.0 & 73.5 & 67.5 & 62.0 \\
& 7 & 89.4 & 81.1 & 73.6 & 67.6 \\
\hline
\end{tabular}
\caption{Percentage of perfect placement for the fragments, with a cut of 0.05.}
\label{tabl:reassembly_fragments}
\end{table}

In Tables \ref{tabl:reassembly_images} and \ref{tabl:reassembly_fragments}, we benchmark the correctness of the reassembly when adding external fragments or removing fragments. We use the network that predicts the position among 9-classes. To build the training set, we set the probability of sampling an outsider fragment to $10\%$.

Table \ref{tabl:reassembly_images} indicates how many images are almost-perfect reassemblies, among our 2000 images test set. Table \ref{tabl:reassembly_fragments} displays the number of well-placed fragments and empty tiles. We emphasize that the random reassembly chance for a 9-pieces jigsaw puzzle with 0 missing and 0 outsiders is upper bounded by $1/8! = 2.4\times10^{-4}$ . This score is to be compared to the 24.7\% almost-perfect reassemblies we obtain in Table \ref{tabl:reassembly_images} and the 22.1\% score with the perfect-only metric. The random scores can be upper-bounded using Equations \ref{eq_cut0} and \ref{eq_cut1}. The almost-perfect metric improves by at least 1.5\% the reassembly score of every configuration of outsider and missing fragments.

Remark that in Table \ref{tabl:reassembly_fragments}, each table cell value is minimized by 11\%, as the center is always well-placed. This minimum is the reason why when we only have to place one fragment (7 missing fragments), we obtain roughly $64\%$ images solved for $89\%$ correct positions. When we do not evaluate the correctness of the empty-tiles positions, we get $64\%$ of well-placed fragments, as a 7-missing fragment puzzle is solved if and only if the only fragment is well-placed.

According to the tables, we obtain best scores either when no fragment is missing or when many fragments are missing. First, when we have all the pieces, we can discriminate similar fragments and select the best one for each location by optimizing the full reassembly. When there are several missing fragments, there is much less information available to assess which one goes where. On the other end of the spectrum, when almost every fragment is missing, the odds we sample the most ambiguous fragment of the image are low.

The outsider results indicate that the more we consider external fragments, the lower the number of almost-perfect reassemblies is, as the number of possible solutions increases. 

To assess how much challenging some puzzles are, we can compare the scores with no outsiders and no missing fragments. We obtain $64.6\%$ of well-placed fragments for only $24.7\%$ of almost-correctly solved puzzles. It means that we often make a few errors in the reassemblies. Indeed, among the not solved puzzle, an average of $55\%$ of the positions predicted are correct.


Thanks to the cutting strategy, we were able to compute reassemblies from a set of 17 fragments quickly. We spend approximatively one hour to construct the graph and apply the shortest path algorithm. Without it, processing more than 3 outsiders fragment could take several months.

\subsubsection{Qualitative reassemblies}

\begin{figure}[htb]
    \centering
    \subfloat[0 missing fragment]{\label{fig:05}{\includegraphics[width=0.23\textwidth]{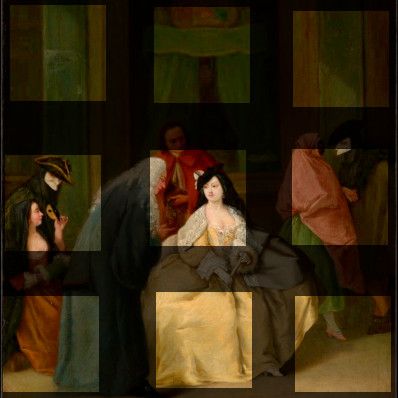}}}\hfill
    \subfloat[5 missing fragments]{\label{fig:07}{\includegraphics[width=0.23\textwidth]{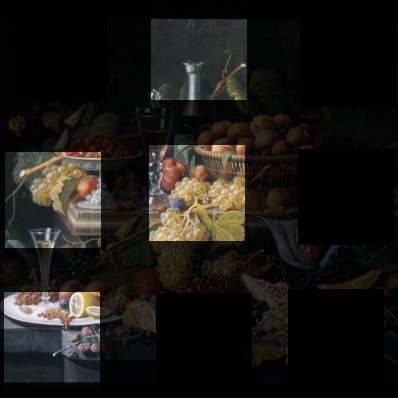}}}
    \caption{Some perfect reassemblies without outsider fragments.}
    \label{fig:reass_without_ext}
\end{figure}

\begin{figure}[htb]
    \centering
    \subfloat[Expected outcome]{\label{fig:06_sol}{\includegraphics[width=0.23\textwidth]{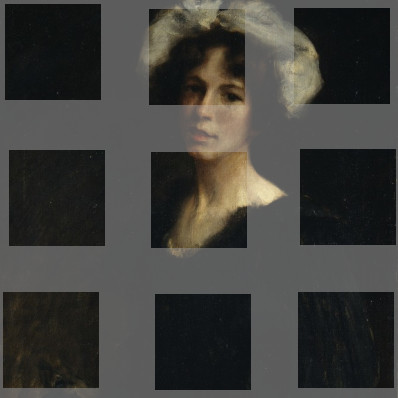}}}\hfill
    \subfloat[Predicted result]{\label{fig:06_res}{\includegraphics[width=0.23\textwidth]{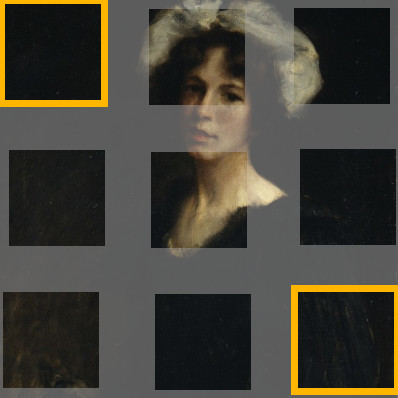}}}
    \caption{An almost perfect reassembly (right). The yellow outline indicates almost-perfectly placed fragments.}
    \label{fig:reass_almost}
\end{figure}

\begin{figure*}[htb]
    \centering
    \subfloat{{\includegraphics[width=0.18\linewidth]{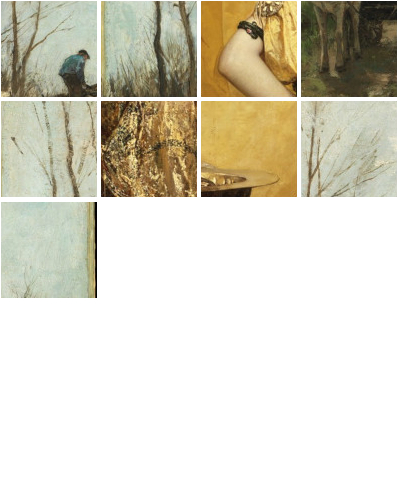}}}\hfill
    \subfloat{{\includegraphics[width=0.18\linewidth]{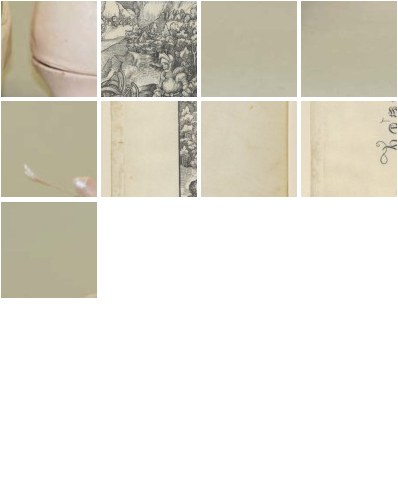}}}\hfill
    \subfloat{{\includegraphics[width=0.18\linewidth]{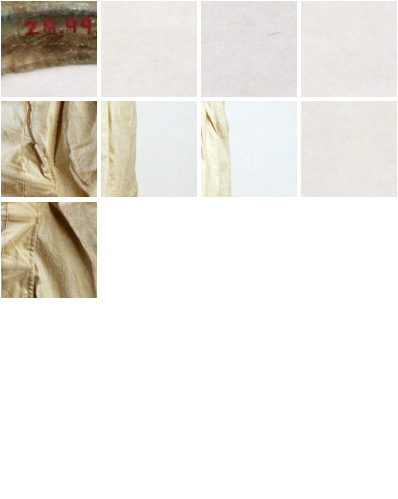}}}\hfill
    \subfloat{{\includegraphics[width=0.18\linewidth]{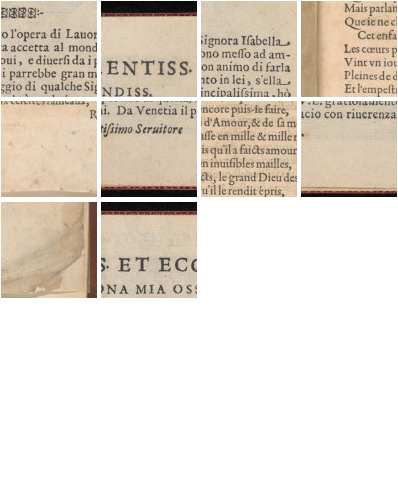}}}\hfill
    \subfloat{{\includegraphics[width=0.18\linewidth]{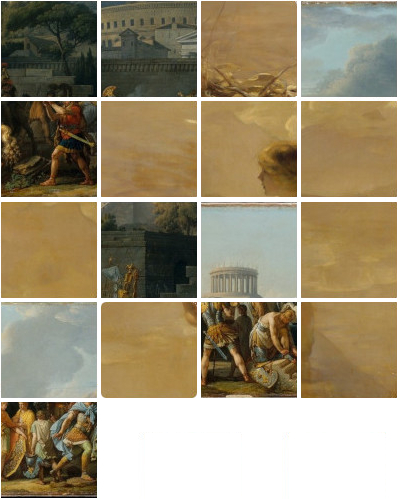}}}
    
    \subfloat{{\includegraphics[width=0.18\linewidth]{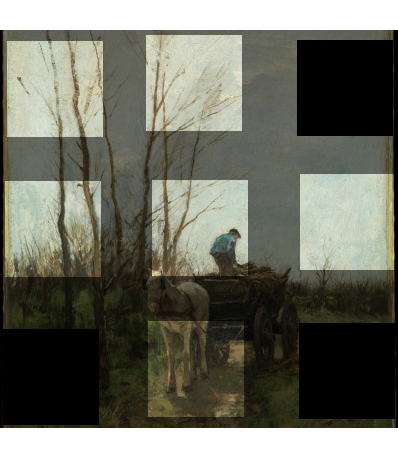}}}\hfill
    \subfloat{{\includegraphics[width=0.18\linewidth]{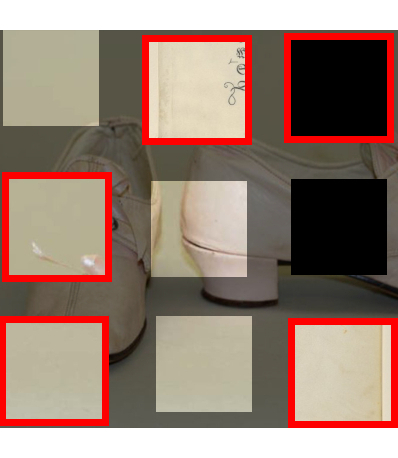}}}\hfill
    \subfloat{{\includegraphics[width=0.18\linewidth]{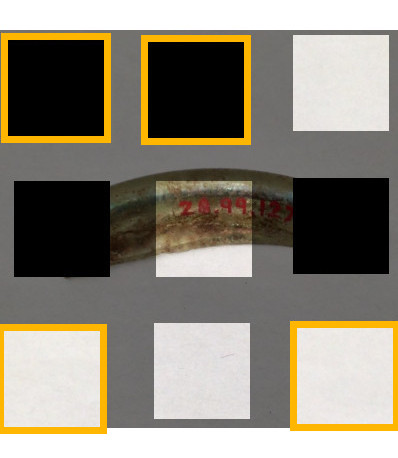}}}\hfill
    \subfloat{{\includegraphics[width=0.18\linewidth]{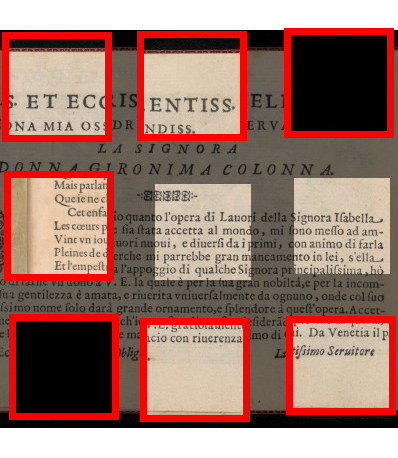}}}\hfill
    \subfloat{{\includegraphics[width=0.18\linewidth]{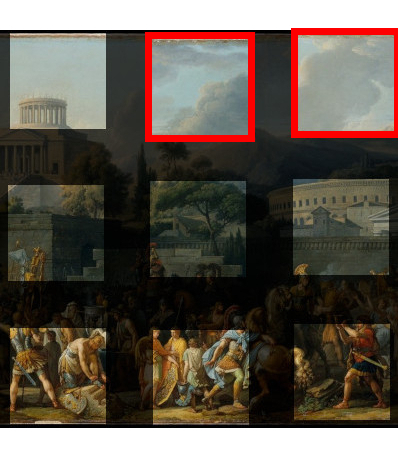}}}
    
    \subfloat{\label{fig:04}{\includegraphics[width=0.18\linewidth]{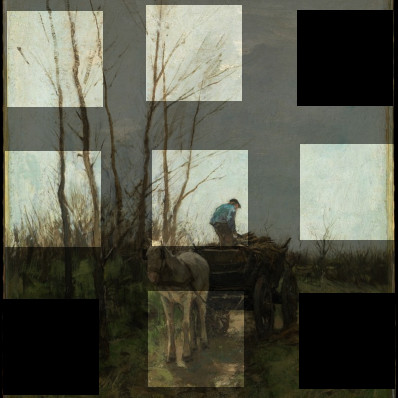}}}\hfill
    \subfloat{\label{fig:01}{\includegraphics[width=0.18\linewidth]{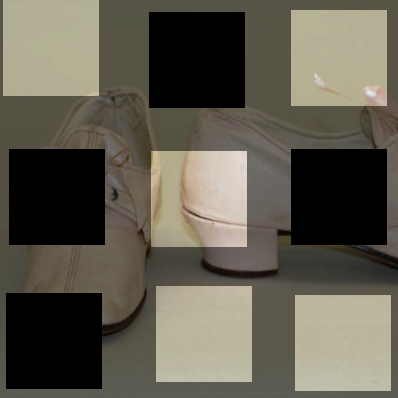}}}\hfill
    \subfloat{\label{fig:02}{\includegraphics[width=0.18\linewidth]{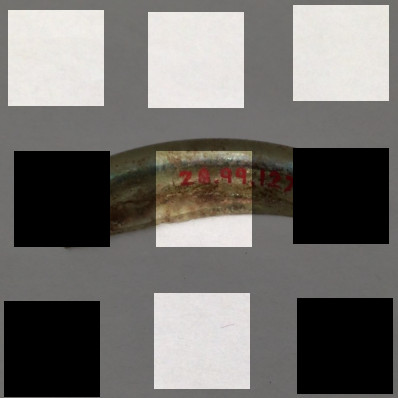}}}\hfill
    \subfloat{\label{fig:03}{\includegraphics[width=0.18\linewidth]{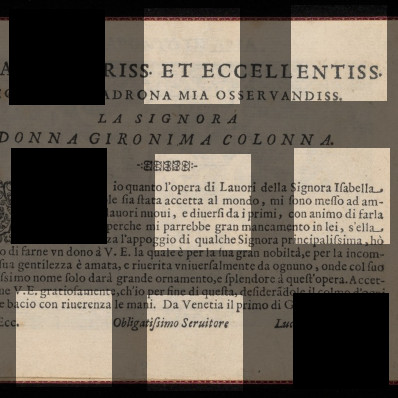}}}\hfill
    \subfloat{\label{fig:08}{\includegraphics[width=0.18\linewidth]{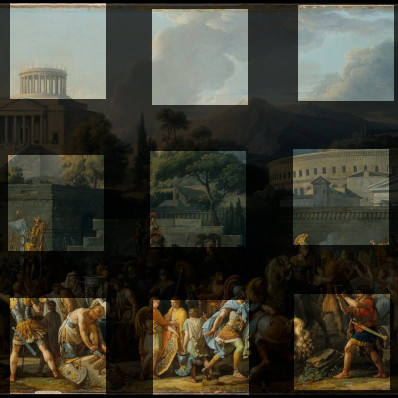}}}
    
    \caption{Various reassemblies with outsider fragments. The first row contains the input fragments. The first (top left) emplacement is reserved to the central fragment. The second row shows the predicted reassemblies. The last row displays the solution. The red outline indicates wrongly placed fragments. The yellow outline shows the almost-perfectly placed fragments.}
    \label{fig:reass_with_add}
\end{figure*}

Figure \ref{fig:reass_without_ext} shows some painting reassemblies in the case there is no outsider fragment. Figures \ref{fig:05} and \ref{fig:07} demonstrate perfect reassemblies with and without missing fragments. In Figure \ref{fig:05}, most of the fragments share color and shape continuity with respects to the central fragment, except the two top corner fragments that display similar probabilities for each of the top position. The rightmost fragment obtains a uniform probability for every position, as its primary color is not part of the central fragment. It is placed correctly because other fragments have been assigned to their correct location thanks to their higher probability.

Figure \ref{fig:07} contains five missing fragments. The remaining fragments are placed correctly because of the composition vacuums and the shape continuity of the table. In most of the painting of the dataset, the painted object is central, which causes the lateral fragments to contain more background space. For instance, the top fragment has details on its bottom and background on its top. Thus, it is correctly placed at the top. Figure \ref{fig:03} also conveys this idea of the expected position of background within a fragment with respect to its assignment.

Figure \ref{fig:reass_almost} demonstrates the utility of the almost-perfect metric. Most of the fragments of this painting are neutral background fragments --- finding which fragment goes where is a random guess. With the new metric, the \ref{fig:06_res} reassembly is considered as correct.

Figure \ref{fig:reass_with_add} contains numerous reassemblies with various cases of outsider fragments. We selected various types of images to highlight the differences of each type (such as paintings and engravings) At first glance, the algorithm tends to replace missing fragments by outsider fragments (Figures \ref{fig:01} and \ref{fig:03}). This observation fits with our analysis of Tables \ref{tabl:reassembly_images} and \ref{tabl:reassembly_fragments}. The switch of missing fragments by outsiders is especially true for clothing, shards, and sculptures backgrounds fragments (Figure \ref{fig:01}). Two other categories of images are prone to be reassembled with outsiders fragments: texts (Figure \ref{fig:03}) and engravings. Conversely, paintings are less exposed to this effect (Figures \ref{fig:04} and \ref{fig:08}), especially when the additional fragments come from non-painting images. When there is no missing fragment (Figure \ref{fig:08}), most of the reassemblies errors are due to misplacing of the inner fragments rather than the replacement of a correct fragment by an additional fragment.  (See Appendix and Figure \ref{fig:texts_reass} for more insight on text puzzles.)

Figure \ref{fig:01} is a typical example of what wrong reassemblies look like: two missing fragments were replaced by similar outsider fragments that contain a mostly-beige background. The shard of Figure \ref{fig:02} is almost-perfectly reassembled, as only background fragments were to be placed.

Figure \ref{fig:03} illustrates the reassembly of a text when another text is the source of the outsider fragments. We obtain poor results (only one fragment is correctly placed), but the spatial coherence of the text is respected. The title is positioned on the top of the image. The fragment that contains the end of the subtitle is at the right of the other title fragment. The end of the text is also placed on the bottom. The italic closing formula is on the right of the other bottom fragment. Finally, the algorithm uses the outsider fragment that contains a left margin at the left of the central fragment. However, the algorithm is not able to distinguish between the French and Italian languages, which suggests the convolutional architecture is not able to learn fine-grain details. This illustrates a limitation of Deepzzle: the input resolution is too small to allow the neural network to capture such details and to produce precise alignment. Deepzzle is intended to solve coarse alignments and thus works best for puzzles with large visual features and sufficient image resolution.

Figure \ref{fig:08} is an example of reassembly with a relatively high number of fragments. The algorithm swapped the cloudy sky fragments. As they are too different pixel-wise, even the almost-perfect metric does not grant the correct reassembly label. Note that to a human eye, the computed reassembly looks realistic with the cloudy sky reversal.

\subsection{Reassembly with unknown centers}

\begin{figure}[!t]
    \centering
    \includegraphics[width=0.75\linewidth]{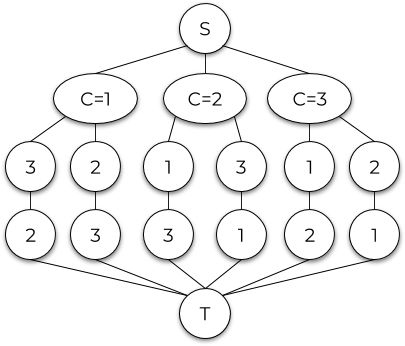}
    \caption{Reassembly graph with various center hypotheses.}
    \label{fig:centers}
\end{figure}

In Table \ref{tabl:dijkstra_vs_greedy}, we compare the accuracies where knowing the central fragments with the case when the central fragment is not known. We do not enable missing nor additional fragments. Note that we use the same 8-classes classifier for each case. In the case where the central fragment is unknown, we alternately use each fragment as the central one and predict the position of every remaining fragment. In the end, we have predictions matrices assuming that each fragment is the central one. We can then build the reassemblies graph, e.g., Figure \ref{fig:centers}.

\begin{table}[!t]
\renewcommand{\arraystretch}{1.3}
\centering
\begin{tabular}{|l|c|c|}
\hline
& \shortstack{Image \\ reassembly} & \shortstack{Fragment \\ position} \\
\hline
Central known & 44.4 & 89.9 \\
Central unknown & 39.2 & 71.1 \\
\hline
\end{tabular}
\caption{Accuracy (\%) for both image reassembly and fragment placement tasks.}
\label{tabl:dijkstra_vs_greedy}
\end{table}

\begin{figure}[!t]
    \centering
    \subfloat[Expected outcome]{\label{fig:09_sol}{\includegraphics[width=0.45\linewidth]{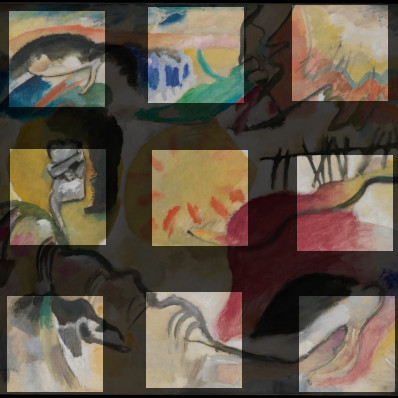}}}\hfill
    \subfloat[Predicted result]{\label{fig:09_res}{\includegraphics[width=0.45\linewidth]{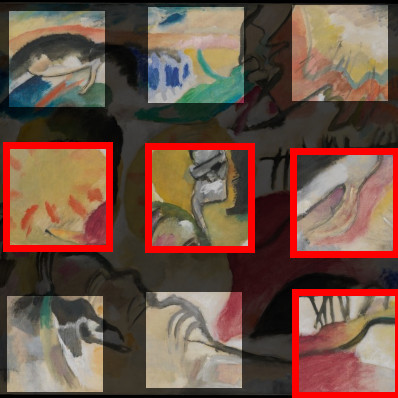}}}\hfill
    \caption{Example of a wrong reassembly with unknown center. The red outline shows the fragments that are misplaced. }
    \label{fig:reass_center}
\end{figure}

The example reassembly shown in Figure \ref{fig:reass_center} raises a particular issue. In this reassembly, the central fragment is unknown. The reassembly algorithm compares the nine graphs of the possible reassemblies given a central fragment and selects the graph that has the smallest shortest path. When the wrong central fragment is chosen, we expect the entire reassembly to be false as every fragment should have been shifted. However, as illustrated by Figure \ref{fig:reass_center}, it occurs that some fragments of the reassembly are well-placed.

As the puzzles are the same with and without knowing the central fragment, it means that the algorithm can choose the graph that contains the correct central fragment. We observe a 5\% drop of the reassembly accuracy, and thus we expect the fragments position accuracy to drop by less than 5\%. Indeed, some reassemblies (Figure \ref{fig:reass_center}) still contains well-placed fragments despite the wrong choice of the central fragment.

\subsection{Reassembly depending on the type of object}

\begin{table}[!t]
    \renewcommand{\arraystretch}{1.3}
    \centering
    \begin{tabular}{|l|c|c|}
        \hline 
        Type of image & \shortstack{Image\\ reassembly} & \shortstack{Fragment\\ position}\\
        \hline
        Artifacts & 38.2 & 70.6\\
        Engravings and texts & 25.5 & 68.0 \\
        Paintings & 12.1 & 56.2 \\
        \hline
        Dataset & 24.7 & 64.6 \\
        \hline
    \end{tabular}
    \caption{Accuracy scores by type of images, without missing nor additional fragments.}
    \label{tabl:type_00}
\end{table}

\begin{table}[!t]
    \renewcommand{\arraystretch}{1.3}
    \centering
    \begin{tabular}{|l|l|c|c|}
        \hline 
        First image & \shortstack{Second \\image} & Reassembly & \shortstack{Fragment \\ position}\\
        \hline
        Artifact & Artifact & 33.0 & 69.2 \\
        Artifact & Engraving & 32.7 & 69.9 \\
        Artifact & Painting & 31.9 & 69.5 \\
        Engraving & Artifact & 22.2 & 67.2 \\
        Engraving & Engraving & 14.1 & 63.3 \\
        Engraving & Painting & 21.4 & 67.0 \\
        Painting & Artifact & 11.5 & 54.9 \\
        Painting & Engraving & 12.5 & 56.7 \\
        Painting & Painting & 11.1 & 54.6 \\
        \hline
        Dataset & Dataset & 17.3 & 60.9 \\
        \hline
    \end{tabular}
    \caption{Accuracy scores by type of images, with two additional fragments and no missing fragments.}
    \label{tabl:type_add}
\end{table}

In Table \ref{tabl:type_00}, we compare the accuracies of almost-perfect image reassembly and fragment position for the three major types of images of our dataset (artifacts, engraving and texts, and painting). We disable the missing and the additional fragments. These scores are to be compared to the average dataset values (last row). The paintings score is surprisingly small on the image reassemblies. It means it is harder to reassemble painting puzzles despite their semantic consistency. As the fragment position score is not as low as we can expect based on the image reassembly score, we conclude that most of the paintings reassemblies only had a very few misplaced fragments. On the contrary, the artifacts score well, primarily because of the almost-perfect metric: the artifacts always have a neutral background (see Figures \ref{fig:01} and \ref{fig:02}).

Table \ref{tabl:type_add} describes the accuracies when there is no missing fragment and two additional fragments extracted from various classes of images. When the image is an artifact picture, we obtain the best results when comparing to another artifact image for the reassembly and when comparing to an engraving for the fragment positioning. As the artifact always comes with a plain background, it is easy for the reassembly algorithm to distinguish the outsider fragments, that rarely have a similar background.

When trying to reassemble an engraving or a text, the best score goes to the artifact additional fragments, closely followed by the painting fragments. The main reason is that engraving or text are like desaturated images, while paintings and artifacts photographs usually come in various colors. Thus, it is more difficult to discriminate the outsider fragments when they come from another engraving or text. It is also why the additional fragments of the engraving score well for the artifacts and the paintings.

\subsection{Reassembly from patchworks}

\begin{figure*}[th]
    \centering
    \subfloat{{\includegraphics[width=0.23\linewidth]{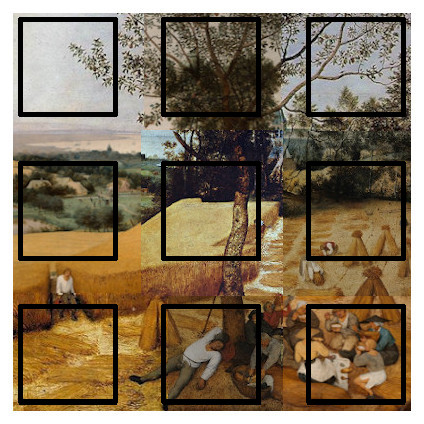}}}\hfill
    \subfloat{{\includegraphics[width=0.23\linewidth]{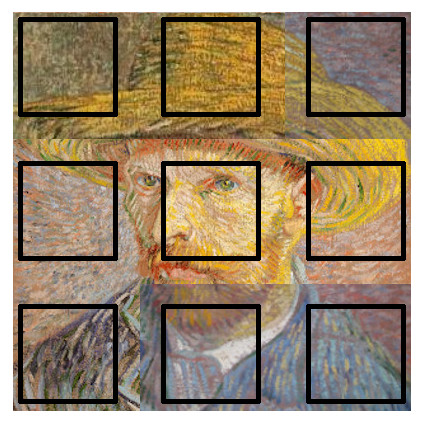}}}\hfill
    \subfloat{{\includegraphics[width=0.23\linewidth]{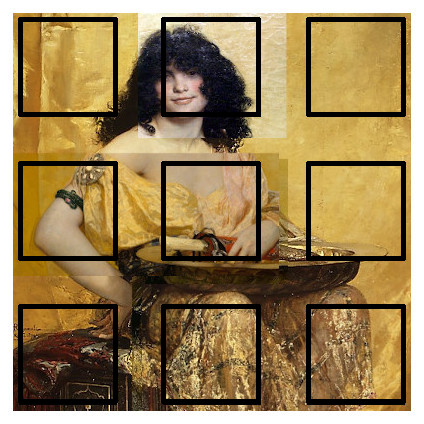}}}\hfill
    \subfloat{{\includegraphics[width=0.23\linewidth]{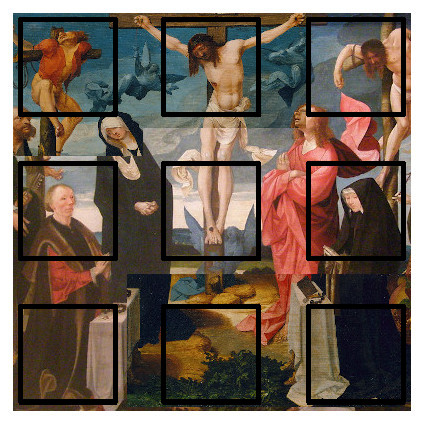}}}

    \subfloat{{\includegraphics[width=0.23\linewidth]{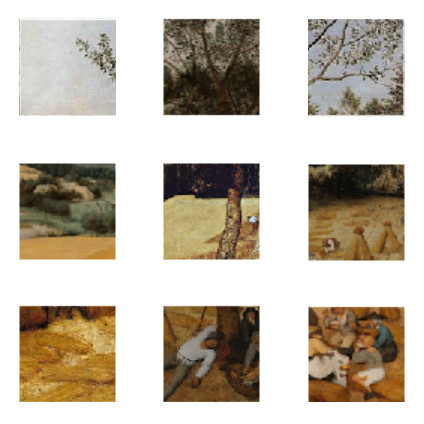}}}\hfill
    \subfloat{{\includegraphics[width=0.23\linewidth]{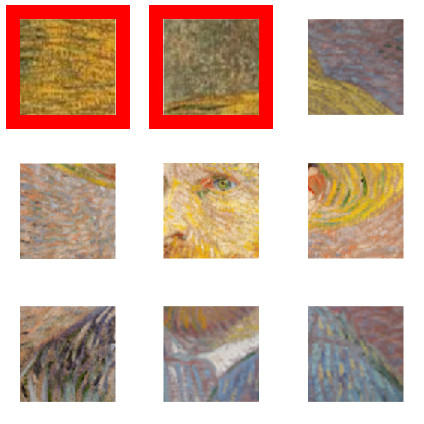}}}\hfill
    \subfloat{{\includegraphics[width=0.23\linewidth]{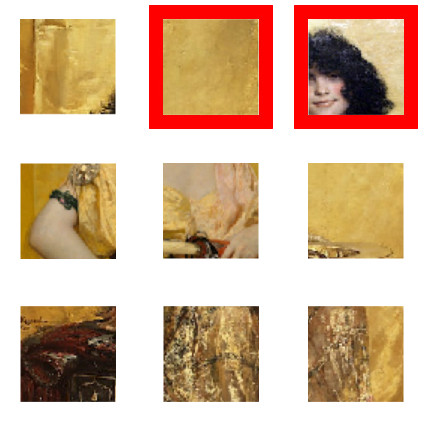}}}\hfill
    \subfloat{{\includegraphics[width=0.23\linewidth]{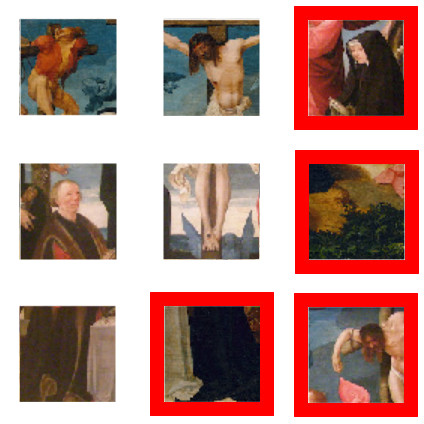}}}
    
    \subfloat{{\includegraphics[width=0.23\linewidth]{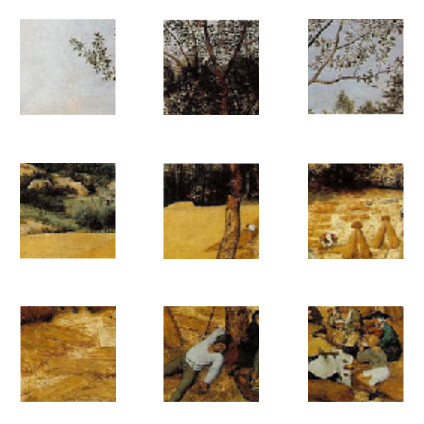}}}\hfill
    \subfloat{{\includegraphics[width=0.23\linewidth]{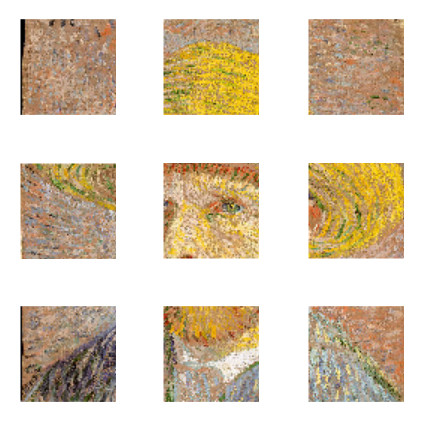}}}\hfill
    \subfloat{{\includegraphics[width=0.23\linewidth]{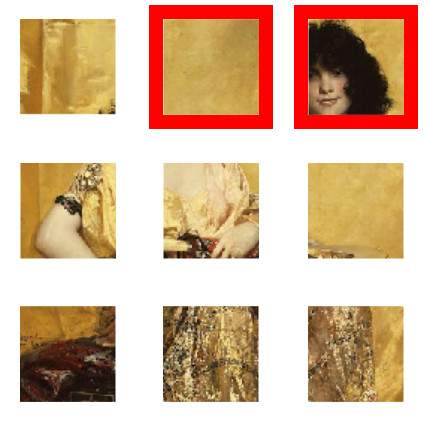}}}\hfill
    \subfloat{{\includegraphics[width=0.23\linewidth]{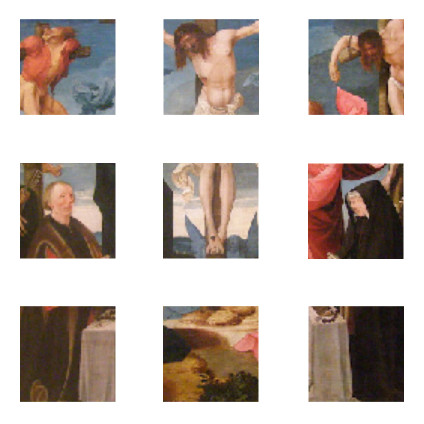}}}

    \caption{Reassemblies from patchwork images. The first row shows the patchwork images from which the fragments were extracted. The second row displays the reassemblies for the patchwork fragments. The third row contains the reassemblies of the MET image (without patchwork). The red outline shows the fragments that are misplaced. }
    \label{fig:patchw}
\end{figure*}

In archaeology, the fragments are photographed independently. The puzzles to solve are made of several tiles coming from different cameras and shooting angles. Merged into one 2D-puzzle, they show slight variations of colors and proportions. We produce 30 patchwork puzzle made from different photographs of some MET paintings, and we solve them (Figure \ref{fig:patchw}). We observed a decrease of 1\% on the number of well-placed fragments, compared to the corresponding MET images. It means that our neural network is not biased by overfitting on the camera parameters.

\section{Conclusion}
This paper deepens the approach of \cite{paumard1} based on five main ideas.
First, we provide a few baselines to evaluate the effectiveness of our method, in comparison with the literature. When necessary, we adapted the methods to make the comparison more meaningful. We obtain better scores on the relative position prediction and on the image reassembly.
Second, we introduce a new metric which takes into account that visually correct solutions are as good as the expected reassembly. Measuring the rate of these almost-perfect reassemblies over the rate of perfect reassemblies leads to an average improvement of 2\% in the reassembly score.
Third, we shorten the computation time of the graph building and solving by 1000, thanks to the cuts in the graphs. This technique leads to a decrease of few tenths in the reassembly and fragments positioning scores. In return, this improvement enables us to solve complete-puzzle with up to 8 additional fragments in less than one hour.
Fourth, we complete the results of \cite{paumard1} with a survey of puzzle-solving in the case of missing and outsiders fragments and a comparison of the cases where the user provide or not the center. We show that the puzzle-solving task with a significant erosion between fragments is challenging, yet we were able to place most of the fragments correctly. Moreover, our last findings indicate that our method is resilient towards photography-related issues.Our method can provide significant aid to archaeologists by cutting down the number of plausible reassemblies.
Last, we analyze the dataset and highlight the specificities of each class of art piece. We demonstrate that the paintings were especially hard to resolve regardless of their spatial consistency.

\appendix
\section{Focus on text reassemblies}

\begin{figure*}[p]
    \centering
    \subfloat{{\includegraphics[width=0.20\linewidth]{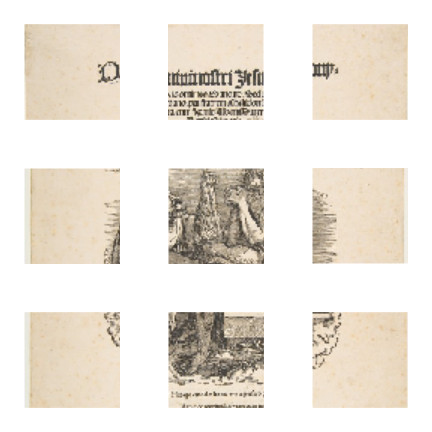}}}\hfill
    \subfloat{{\includegraphics[width=0.20\linewidth]{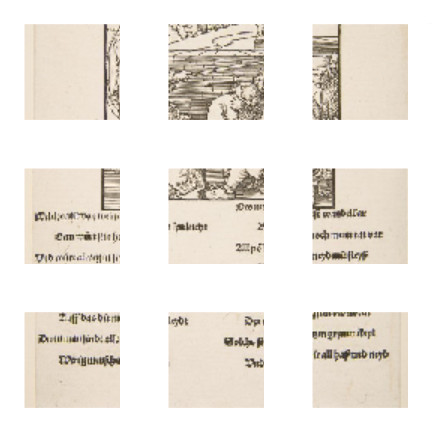}}}\hfill
    \subfloat{{\includegraphics[width=0.20\linewidth]{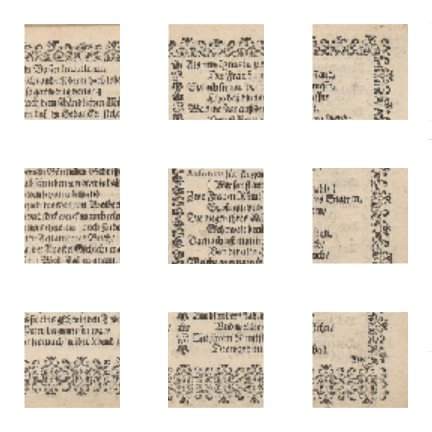}}}\hfill
    \subfloat{{\includegraphics[width=0.20\linewidth]{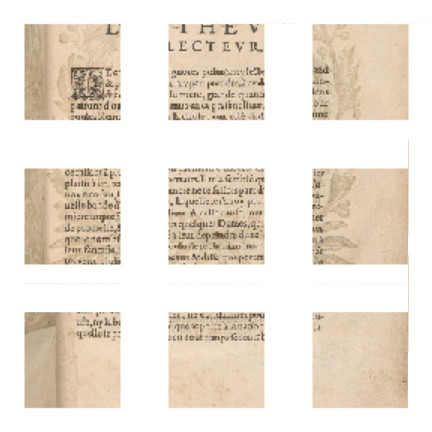}}}

    \subfloat[]{\label{fig:11a}{\includegraphics[width=0.20\linewidth]{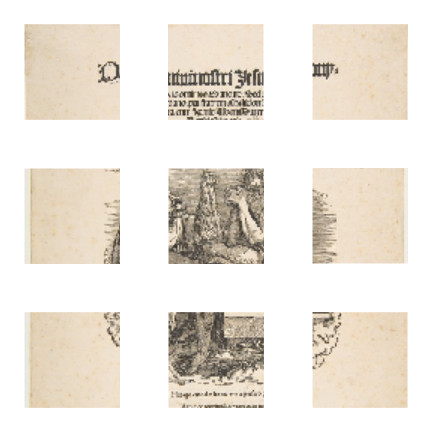}}}\hfill
    \subfloat[]{\label{fig:11b}{\includegraphics[width=0.20\linewidth]{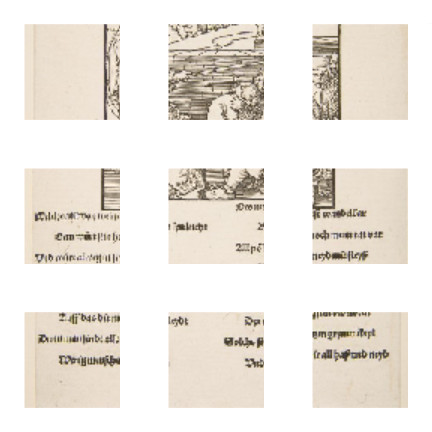}}}\hfill
    \subfloat[]{\label{fig:11c}{\includegraphics[width=0.20\linewidth]{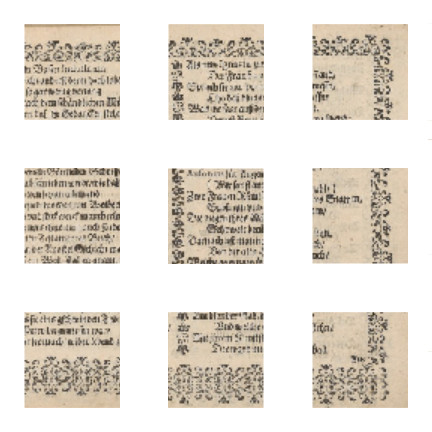}}}\hfill
    \subfloat[]{\label{fig:11d}{\includegraphics[width=0.20\linewidth]{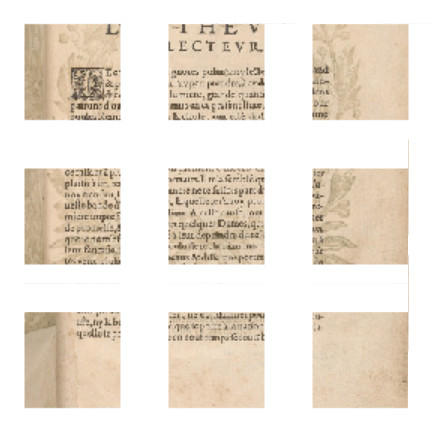}}}

    \subfloat{{\includegraphics[width=0.20\linewidth]{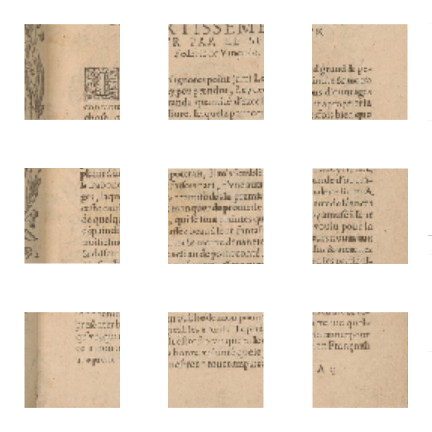}}}\hfill
    \subfloat{{\includegraphics[width=0.20\linewidth]{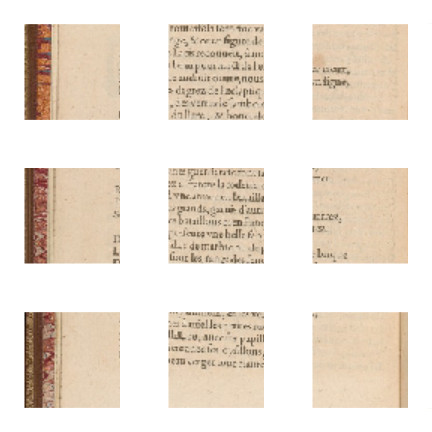}}}\hfill
    \subfloat{{\includegraphics[width=0.20\linewidth]{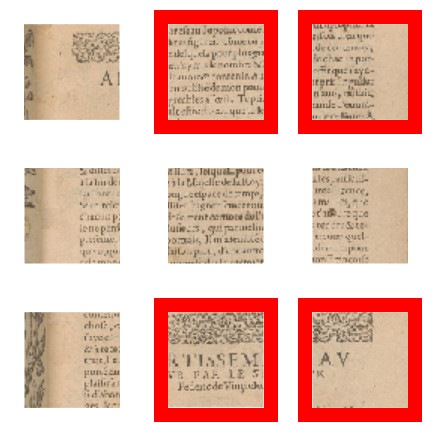}}}\hfill
    \subfloat{{\includegraphics[width=0.20\linewidth]{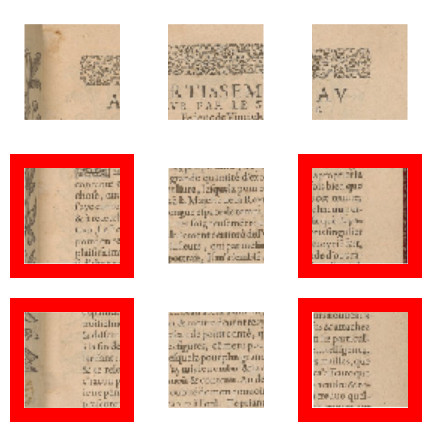}}}

    \subfloat[]{\label{fig:11e}{\includegraphics[width=0.20\linewidth]{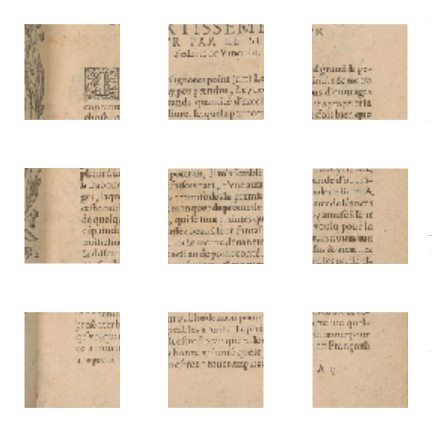}}}\hfill
    \subfloat[]{\label{fig:11f}{\includegraphics[width=0.20\linewidth]{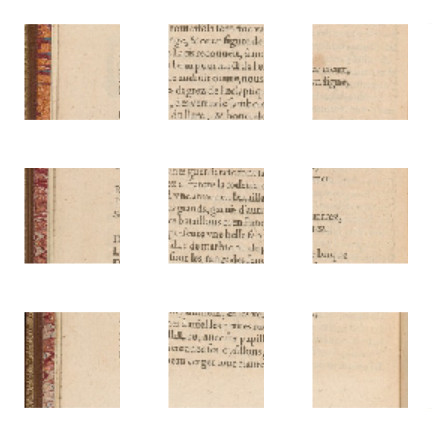}}}\hfill
    \subfloat[]{\label{fig:11g}{\includegraphics[width=0.20\linewidth]{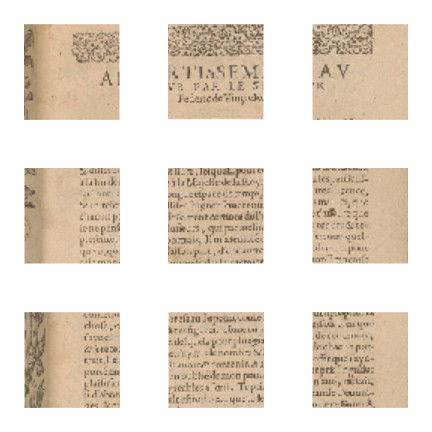}}}\hfill
    \subfloat[]{\label{fig:11h}{\includegraphics[width=0.20\linewidth]{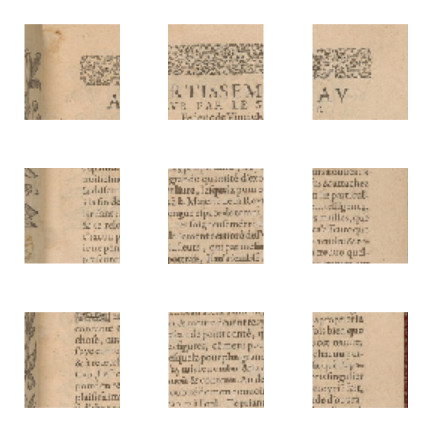}}}

    \subfloat{{\includegraphics[width=0.20\linewidth]{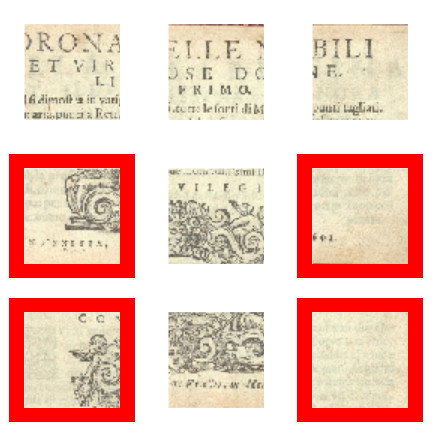}}}\hfill
    \subfloat{{\includegraphics[width=0.20\linewidth]{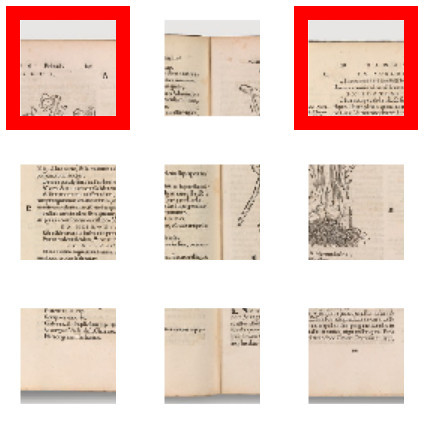}}}\hfill
    \subfloat{{\includegraphics[width=0.20\linewidth]{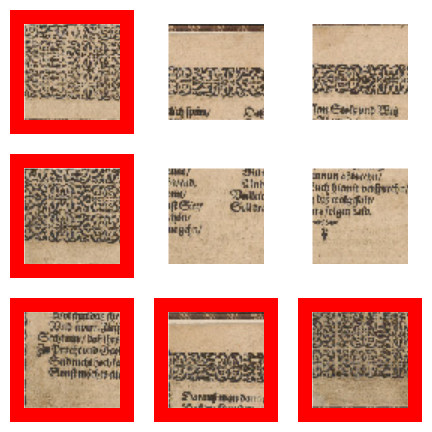}}}\hfill
    \subfloat{{\includegraphics[width=0.20\linewidth]{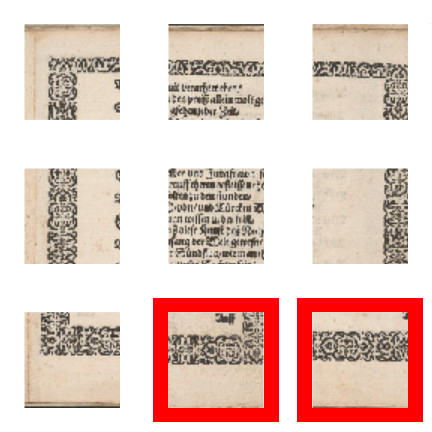}}}

    \subfloat[]{\label{fig:11i}{\includegraphics[width=0.20\linewidth]{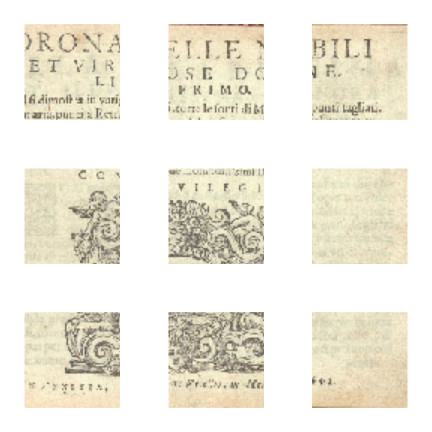}}}\hfill
    \subfloat[]{\label{fig:11j}{\includegraphics[width=0.20\linewidth]{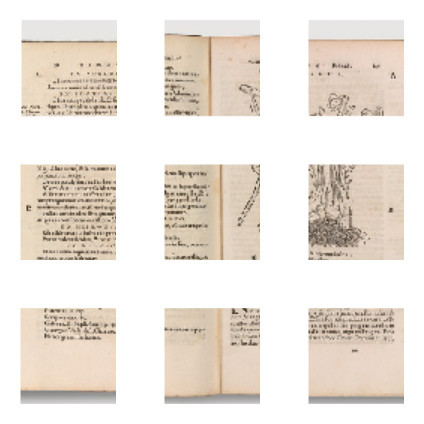}}}\hfill
    \subfloat[]{\label{fig:11k}{\includegraphics[width=0.20\linewidth]{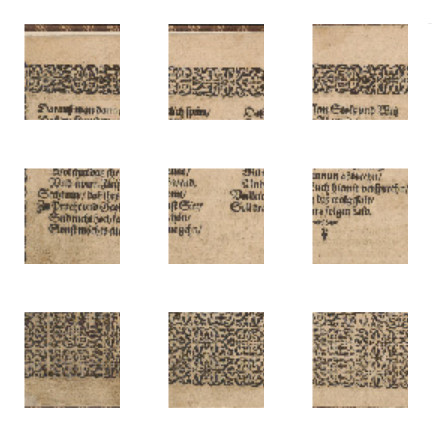}}}\hfill
    \subfloat[]{\label{fig:11l}{\includegraphics[width=0.20\linewidth]{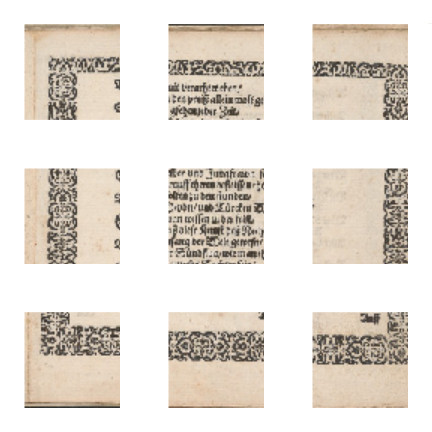}}}
    
    \caption{Predicted reassemblies (odd rows) and their solutions (even rows) for texts. The red outline shows the fragments that are misplaced. }
    \label{fig:texts_reass}
\end{figure*}

In this appendix, we analyze reassemblies of texts. We aim to gain insights on the patterns that are used by the neural network to make its predictions.
We select thirty text pictures from the MET dataset. We obtained 24\% of perfect reassemblies and 68\% of well-placed fragments, which is consistent with Table \ref{tabl:type_00}.

We append numerous text reassembly images in Figure \ref{fig:texts_reass}. Figure \ref{fig:11a} is a perfect example of confident reassembly: most of fragments positions are predicted with a confidence score superior to 70\%. In this image, the only fragments whose correct class is not the most confident one are the upper right fragments (24\% for the upper right position, against 36\% for the bottom left position).

The central fragments of Figures \ref{fig:11b} and \ref{fig:11c} contains clues about how to solve the puzzle. Looking at the central fragment of Figure \ref{fig:11b}, we have an image on top that probably stretches out over the top fragments. We also have text at the bottom left and at the bottom right of the central fragment, with a space between them. By extending all of these structures, one can easily solve the puzzle. Each relative prediction is correctly predicted with confidence over 50\%.

Figures \ref{fig:11d}, \ref{fig:11e}, \ref{fig:11f}, \ref{fig:11g} and \ref{fig:11h} shows a central text fragment. In Figures \ref{fig:11d} and \ref{fig:11e}, the lateral fragments contains text and margin in the four directions, and are well reassembled. Figure \ref{fig:11f} is perfectly reassembled by chance, as the left and right fragments vertical position display very close classifications scores. Figures \ref{fig:11g} and \ref{fig:11h} contains the same puzzle, with a vertical shift.

Figure \ref{fig:11g} is interesting, as most title fragments were placed at the bottom of the puzzle. These two fragments are similar text fragments, because there is no space between the top of the fragments and the horizontal ornamentation. We suppose this similarity is the cause of the misplacement. On the contrary, the upper left fragment contains space before the frieze: then it cannot continue the text. The correct position of the title in Figure \ref{fig:11h} supports this idea. Looking to first predicted class scores in Figures \ref{fig:11g} and \ref{fig:11h}, we observe a strong vertical arrangement with close position scores (with a difference lower than 5\% between the vertical positions scores).

On Figure \ref{fig:11i}, the reassembly display mistakes on similar fragments. On Figure \ref{fig:11j}, the position of the fragments that display the bookbinding are correctly predicted (the first classes are at 75\% and 77\% respectively). The fragments are placed correctly in the horizontal axis, but the right and left upper fragment are unluckily swapped (their scores for the various top positions classes are around 30\%).

Figures \ref{fig:11k} and \ref{fig:11l} illustrates that texts and ornaments are not distinguished by the neural network. When the fragments are well placed, it is because of its white space.

In summary, text reassembly primarily uses borders, margins, and frames. They often identify the fragments being part of the same column (and, more rarely, the fragments from the same row).

\section*{Acknowledgments}
The authors would like to thank the Fondation des Sciences du Patrimoine (ANR EUR-17-EURE-0021).

\bibliographystyle{plain}
\bibliography{sources}

\end{document}